\documentclass[11pt]{article}
\usepackage[final]{acl}

\usepackage{xcolor}
\usepackage{float}
\usepackage{times}
\usepackage{latexsym} 
\usepackage[T1]{fontenc}
\usepackage[utf8]{inputenc}
\usepackage{microtype}
\usepackage{inconsolata}
\usepackage{graphicx}
\usepackage{booktabs}
\usepackage{amsmath}
\usepackage{multirow}
\usepackage{xcolor}
\usepackage{url}
\usepackage{placeins}
\usepackage{comment}

\newcommand{\same}[1]{{\color{gray}\fontsize{7pt}{7pt}\selectfont\hspace{1pt}(\pm0.000)}}
\usepackage[most]{tcolorbox}
\tcbuselibrary{breakable}
\usepackage{fvextra}
\usepackage{kotex}

\newtcolorbox{PromptBox}[1]{
  breakable,
  colback=gray!5!white,
  colframe=gray!55!black,
  boxrule=0.5pt,
  left=1mm,
  right=1mm,
  top=1mm,
  bottom=1mm,
  title={\small\bfseries #1}
}

\DefineVerbatimEnvironment{PromptVerbatim}{Verbatim}{
  breaklines=true,
  breakanywhere=false,
  breakautoindent=false,
  breakindent=0pt,
  breaksymbolleft={},
  breaksymbolright={},
  fontsize=\footnotesize
}

\title{Beyond Consensus: Downward Bias and Role Asymmetry in Multi-Agent LLM Judges for Subjective Evaluation}

\author{
 \textbf{Minsoo Song\textsuperscript{1}},
 \textbf{Chanwoo Kim\textsuperscript{1}},
 \textbf{Sugyeong Eo\textsuperscript{2, \textdagger}},
 \textbf{Chanjun Park\textsuperscript{1,\textdagger}}
\\
\\
 \textsuperscript{1}Soongsil University \textsuperscript{2} Yonsei University 
\\
 \texttt{\{ecoses042, kcw4776\}@soongsil.ac.kr, s.eo@yonsei.ac.kr,} 
 \\
 \texttt{chanjun.park@ssu.ac.kr}
\\
}
\begin{document}
\maketitle

\begingroup
\renewcommand{\thefootnote}{\textdagger}
\footnotetext{Corresponding authors.}
\endgroup

\begin{abstract}
Multi-Agent Debate (MAD) has been widely adopted to improve LLM-based evaluation by prompting multiple agents to negotiate and reach a consensus. However, for subjective rubric-based scoring, inter-agent agreement does not guarantee alignment with human judgments. In this paper, we compare a single-judge baseline against a consensus-based MAD protocol on subjective evaluation tasks and design three ablations to isolate the impact of role prompting, multi-round interaction, and explicit score sharing. Evaluations across six LLMs show that the single-judge baseline  achieves the strongest human alignment on average across six judge models, whereas MAD shows degradation in human alignment on both tasks. Our ablations demonstrate that this performance drop stems primarily from asymmetric role prompting rather than the interaction itself. Specifically, assigning a strict judge role introduces a systematic downward bias that the consensus process fails to correct. The central finding is that this bias reflects strict-stance dominance beyond averaging: the consensus score falls well beyond the arithmetic midpoint of the standalone strict and lenient conditions, rather than averaging them out. Removing role asymmetry (Symmetric MAD) largely recovers baseline performance, while masking peer scores widens inter-agent disagreement on average and worsens average human alignment. These findings demonstrate that multi-agent consensus can enforce artificial agreement at the expense of true human alignment, revealing a structural limitation in consensus-style, role-specialized MAD protocols for subjective scoring.
\end{abstract}

\section{Introduction}
% ── Figure 1전체적인 실험 흐름) ──────────────
\begin{figure}[h]
  \centering
  \includegraphics[width=\linewidth]{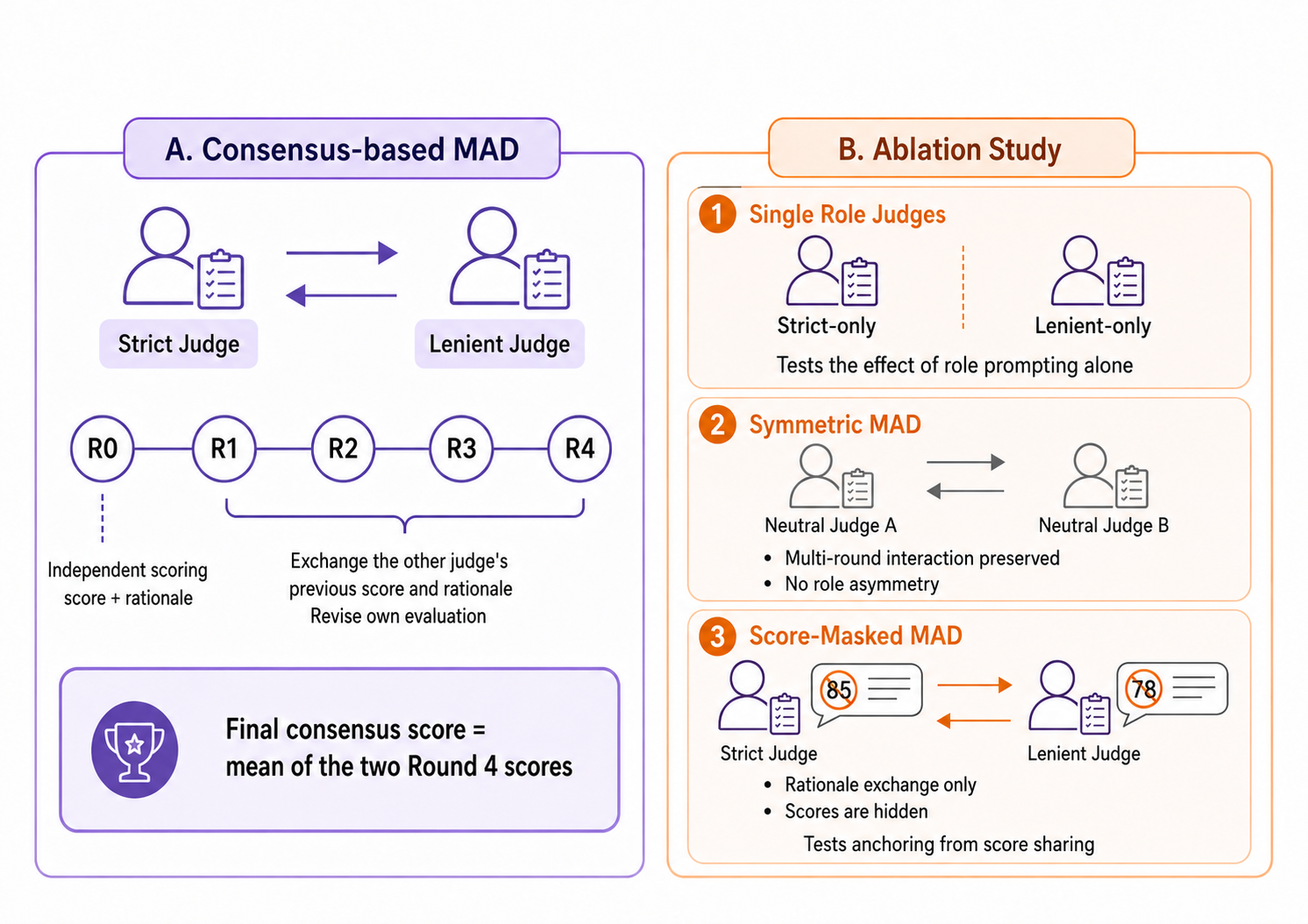}
  \caption{Overview of the experimental design. We compare a single-judge baseline with consensus-based MAD and three ablations that isolate role prompting, debate interaction, and score sharing, using RMSE and Spearman correlation against human reference scores.}
  \label{fig:overview}
\end{figure}
LLM-as-a-Judge~(LLM judge) has become a widely adopted paradigm for evaluating open-ended language generation tasks. Prior studies have shown that LLMs can provide scalable and rubric-aware evaluations for summarization, dialogue, and long-form responses, often showing stronger correlations with human judgments than traditional metrics~\cite{liu2023g,zhu2025judgelm,kim2024prometheus}. However, LLM judges are also known to exhibit systematic biases, such as self-preference, position bias, and unfair evaluation behavior~\cite{wataoka2024self,shi2025judging,wang2024large}.

In parallel, Multi-Agent Debate~(MAD) has been proposed as a way to improve the reasoning and factuality of LLM outputs by allowing multiple agents to exchange and revise their answers~\cite{du2024improving,liang2024encouraging}. Recent work has extended this idea to LLM-based evaluation, introducing multi-agent or debate-based judges for open-ended text evaluation, evaluator meta-evaluation, and machine translation evaluation~\cite{chan2024chateval,li2024mateval,chern2024can,feng2025m,chen2025multi}.
Such applications are supported by evidence that MAD can be effective in reasoning-oriented tasks where answers are relatively verifiable, such as factual question answering or mathematical reasoning~\cite{liang2024encouraging, zhang2025debate4math}. However, these studies do not systematically evaluate whether multi-agent judging improves alignment with human reference scores in criterion-level subjective scoring tasks.

To this end, we investigate whether consensus-based MAD improves or degrades human alignment in subjective evaluation, and, if degradation is observed, what mechanisms may explain it. We treat numerical anchoring and role-prompt asymmetry as two competing hypotheses for this degradation rather than as an established mechanism, and we design ablations to test each hypothesis directly. Under the anchoring hypothesis, score sharing induces anchoring effects: prior work on judgment under uncertainty shows that exposure to numerical values biases subsequent judgments toward observed anchors~\cite{tversky1974judgment,mussweiler2000numeric}, so agents observing both textual rationales and numerical scores from other agents may converge toward shared score ranges, compressing score distributions and reducing discriminability. Under the role-prompt asymmetry hypothesis, assigning agents asymmetric evaluative stances, such as strict and lenient judge roles, allows one stance to dominate the deliberation process and weaken the reliability of the final consensus independent of whether numerical scores are shared. Score-Masked MAD tests the anchoring hypothesis by removing numerical score sharing while preserving role asymmetry, while Symmetric MAD tests the role-prompt asymmetry hypothesis by removing role asymmetry while preserving multi-round interaction.

To investigate further, we compare single-agent and Consensus-based MAD judges against human reference scores, using ablations that isolate score sharing and asymmetric role prompting. Our experiments show that consensus-style multi-agent judging degrades alignment with human reference scores, and our ablation analysis suggests that asymmetric role prompting is a primary driver of this degradation.
\section{Methodology}
\label{sec:methodology}

We study a compact consensus-style multi-agent judging design for subjective evaluation tasks. To isolate and examine the core mechanisms shared by prior multi-agent debate and agent-as-judge studies~\cite{du2024improving, chan2024chateval, li2024mateval}, we deliberately instantiate a minimal MAD protocol with role-specialized agents, iterative exchange, and final aggregation, rather than proposing a new framework.
This design allows us to test whether a consensus-oriented interaction, which has been used to improve reasoning or open-ended evaluation, also improves alignment with human ratings in subjective scoring. Figure~\ref{fig:overview} summarizes the experimental design. Across all protocols, we keep the task input, target text, rubric, scoring scale, and judge model fixed. The full prompts are provided in Appendix~\ref{app:prompts}.

\subsection{Judge Protocols}
\label{sec:judge_protocols}

\paragraph{Single Judge}
The single-judge condition serves as the baseline. 
A single LLM judge receives the task instruction, rubric, and target text, and independently assigns criterion-level scores. 
This condition involves no role specialization, inter-agent interaction, or score aggregation.

\paragraph{Consensus-based MAD}
Our main multi-agent condition consists of two role-specialized agents: a Strict Judge and a Lenient Judge.
We use this strict–lenient pair as a minimal form of evaluative diversity: subjective scoring often involves a tension between penalizing weaknesses and recognizing plausible strengths. At Round 0, both agents independently evaluate the target text. In each subsequent round, the strict and lenient agents observe each other's previous score and rationale, then revise or maintain their own judgments. The final prediction is the arithmetic mean of the two agents' final-round scores. In our experiments, we conducted this process up to Round 4.
\subsection{Ablation Conditions}
\label{sec:ablation_conditions}

We introduce three ablation conditions to decompose the consensus-style design into role prompting, multi-round interaction, and numerical score sharing.

\paragraph{Strict Role Judge}
The Strict Role Judge applies the same strict scoring prompt used in the role-specialized consensus judge, but removes inter-agent interaction. 
This condition tests whether the strict role prompt alone changes score levels or human alignment. 
Comparing this condition with the single-judge baseline isolates the effect of role framing, while comparing it with the Consensus-based MAD shows whether interaction amplifies or moderates the effect of the strict role.

\paragraph{Symmetric MAD}
Symmetric MAD preserves the same multi-round interaction structure as the main consensus protocol but removes role specialization. 
Both agents use the same neutral judge prompt and exchange previous-round evaluations. 
This condition tests whether interaction alone explains the observed behavior, or whether asymmetric role prompting is necessary for the alignment changes observed in the main protocol.

\paragraph{Score-Masked MAD}
Score-Masked MAD preserves the strict--lenient role structure and rationale exchange, but removes explicit numerical score sharing from inter-agent messages.
Agents observe the other agent's previous rationale excluding any numerical values.
This condition tests whether numerical scores act as anchoring signals during deliberation, since prior work shows that exposure to numerical values can bias subsequent judgments under uncertainty~\cite{tversky1974judgment,mussweiler2000numeric}.
\section{Experimental Setup}
\label{sec:experimental_setup}

\subsection{Tasks and Datasets}
\label{sec:tasks_datasets}

We evaluate judge protocols on two subjective evaluation tasks that provide criterion-level human reference scores for rubric-based evaluation, enabling direct measurement of judge-to-human alignment. Detailed dataset settings are provided in Appendix~\ref{app:dataset_details}.

We deliberately diversify language, rubric structure, and genre across the two tasks, while standardizing the measurement protocol: both tasks use a 1--5 scoring scale, criterion-level human reference scores, and an identical grid of judge protocols and judge models. This design lets us test whether the effects of role-specialized consensus generalize across domains that differ substantially in surface characteristics but are evaluated under the same experimental grid.

\paragraph{Korean Essay Scoring}
For this task, we use the National Institute of Korean Language writing assessment corpus\footnote{\url{https://kli.korean.go.kr/corpus/request/corpusRegist.do}}. We sample 600 essays from six topics. Each essay is evaluated on three criteria: content, organization, and expression. The average of two human evaluators is leveraged as the reference score.

\paragraph{Summarization Evaluation}
We use SummEval~\cite{fabbri2020summeval}, which provides human ratings for system-generated summaries across four criteria: coherence, consistency, fluency, and relevance. 
We use summaries generated by seven systems for 100 source articles, yielding 700 system-summary instances. 
Each summary is evaluated at the criterion level, and the human rating is used as the reference score.

One prompt-level asymmetry exists between the two single-judge baselines: the SummEval single-judge prompt includes the phrase \textit{strict and consistent evaluator}, a directional cue absent from the Korean Essay single-judge prompt. Because our own ablations show that a strict framing systematically lowers human alignment relative to a neutral prompt (Table~\ref{tab:main}), this asymmetry, if anything, works against our central claim: it nudges the SummEval single-judge baseline toward the same strict framing that degrades alignment under Consensus-based MAD, which would narrow rather than inflate the single-judge advantage we report. We therefore consider the direction of this asymmetry conservative with respect to our conclusions, although we do not directly quantify its magnitude.

\subsection{Models and Metrics}
\label{sec:models_metrics}

We use six LLMs as judge models: GPT-4o-mini\footnote{\url{https://developers.openai.com/api/docs/models/gpt-4o-mini}}, Gemma-3-4B-IT, Gemma-3-12B-IT, and Gemma-3-27B-IT~\cite{gemmateam2025gemma3technicalreport}, Qwen3.5-9B, and Qwen3.5-27B~\cite{qwen3.5}. We select models spanning three families and three capacity levels (4B--27B) to assess whether findings hold across model sizes and architectures. All models are accessed through OpenRouter\footnote{\url{https://openrouter.ai/}} with temperature setting of 0. Within each run, the same model performs all judge roles. Additional inference details are provided in Appendix~\ref{app:model_inference_details}. For measuring human alignment, we use RMSE and Spearman correlation. RMSE measures absolute agreement with human reference scores, while Spearman correlation measures whether the protocol preserves the relative ranking of instances.

%% ============================================================
%%  §4  Results
%% ============================================================

\section{Results}
\label{sec:results}
% ── Table 1: RMSE + Spearman 요약 ─────────────────────────────
\begin{table}[t]
  \centering
  \small
  \resizebox{\columnwidth}{!}{
  \begin{tabular}{lcccc}
    \toprule
    & \multicolumn{2}{c}{\textbf{Korean Essay}}
    & \multicolumn{2}{c}{\textbf{SummEval}} \\
    \cmidrule(lr){2-3}\cmidrule(lr){4-5}
    \textbf{Protocol} & RMSE $\downarrow$ & $\rho$ $\uparrow$ & RMSE $\downarrow$ & $\rho$ $\uparrow$ \\
    \midrule
    Single Judge        & \textbf{0.644} & \textbf{0.504}
                        & \textbf{0.682} & \textbf{0.458} \\
    \midrule
    Strict Role Judge   & 0.973 & 0.499 & 0.884 & 0.447 \\
    Symmetric MAD       & 0.733 & 0.451 & 0.687 & 0.447 \\
    Consensus-based MAD & 0.935 & 0.424 & 0.813 & 0.409 \\
    Score-Masked MAD    & 1.040 & 0.399 & 0.994 & 0.368 \\
    \bottomrule
  \end{tabular}}
\caption{Human alignment across judge protocols, averaged over six judge models. The single-judge baseline achieves the strongest average alignment on two datasets.}
  \label{tab:main}
\end{table}

\begin{figure}[t]
  \centering
  \includegraphics[width=\linewidth]{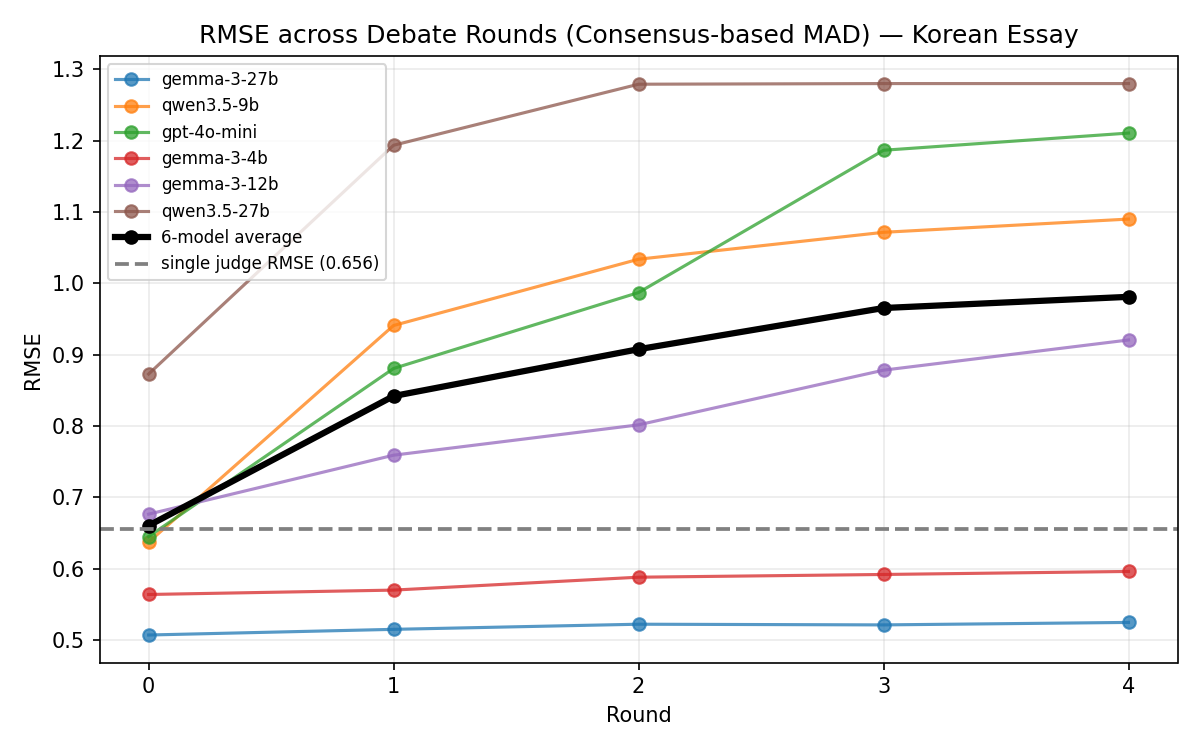}
\caption{Round-wise RMSE under Consensus-based MAD on the Korean Essay scoring task. 
Each point traces the protocol-level prediction after each exchange round. 
The upward trend after early interaction suggests that iterative consensus does not necessarily move judge predictions closer to human reference scores.}
  \label{fig:round_alignment}
\end{figure}
\begin{table}[t]
  \centering
  \small
  \setlength{\tabcolsep}{4pt}
  \begin{tabular}{lcc}
    \toprule
    \textbf{Protocol} & \textbf{Korean Essay} & \textbf{SummEval} \\
    \midrule
    Reference           & \textbf{3.486} & \textbf{4.285} \\
    \midrule
    Single Judge        & 3.345 & 3.935 \\
    \midrule
    Strict Role Judge   & 2.733 & 3.672 \\
    Symmetric MAD       & 3.249 & 3.964 \\
    Consensus-based MAD & 2.900 & 3.831 \\
    Score-Masked MAD    & 2.767 & 3.667 \\
    \bottomrule
  \end{tabular}
\caption{Mean predicted score per protocol, averaged over all evaluation instances and six judge models.
Role-asymmetric protocols show a consistent downward shift relative to the single-judge baseline and human reference. Appendix~\ref{app:score_masking} provides detailed analysis.}
  \label{tab:meanscore}
\end{table}

\subsection{Human Alignment Across Protocols}
\label{sec:human_alignment}

Table~\ref{tab:main} reports RMSE and Spearman correlation averaged over six judge models.
The single-judge baseline achieves the best alignment on both tasks in both RMSE and Spearman correlation.
Consensus-based MAD increases RMSE to 0.935 and 0.813 while Spearman drops to 0.424 and 0.409,
indicating degradation in both absolute agreement and ranking preservation.
As illustrated in Figure~\ref{fig:round_alignment}, RMSE generally increases after the initial exchange. This indicates that subsequent rounds fail to correct the degradation and instead compound it. Detailed per-model results and gap trajectories can be found in Appendix~\ref{app:Model Specific Analysis}. Significance tests for this degradation are reported in Appendix~\ref{app:significance}.

\subsection{Role Asymmetry}
\label{sec:role_specialization}

Table~\ref{tab:meanscore} reports mean predicted scores, which indicate the
direction of score shift rather than human alignment itself. Consensus-based
MAD produces lower mean scores than both the single-judge baseline and the
human reference on both tasks. On Korean Essay, the mean score is 2.900 under
Consensus-based MAD, compared with 3.345 under Single Judge and 3.486 for the
human reference. On SummEval, the corresponding values are 3.831, 3.935, and
4.285. This shows that the strict--lenient consensus protocol shifts the
average prediction downward relative to the human reference.

This downward shift is most pronounced in the Strict Role Judge condition.
Its mean scores are 2.733 on Korean Essay and 3.672 on SummEval, both below
the Consensus-based MAD means. The same condition also yields higher RMSE
than Consensus-based MAD (Essay: 0.973 vs.\ 0.935; SummEval: 0.884 vs.\ 0.813),
indicating that the strict role prompt alone moves predictions away from
the human reference. Interaction in Consensus-based MAD partially moderates
this strict-role shift, but does not restore single-judge alignment. Dominance analysis results are reported in Appendix~\ref{app:lenient_role}.

By contrast, Symmetric MAD remains much closer to the single-judge baseline.
Its RMSE is 0.733 on Korean Essay and 0.687 on SummEval, compared with
0.644 and 0.682 for the Single Judge. Its mean scores are also closer to the
single-judge means (Essay: 3.249 vs.\ 3.345; SummEval: 3.964 vs.\ 3.935).
Together, these results suggest that the main source of alignment loss is not
multi-round interaction itself, but asymmetric role prompting. Lenient Role Judge results are reported in Appendix~\ref{app:lenient_role}.

\subsection{Cross-Model Generalization}
\label{sec:cross_model}

\begin{table}[h]
\centering
\small
\setlength{\tabcolsep}{3pt}
\resizebox{\columnwidth}{!}{
\begin{tabular}{lcc}
\toprule
\textbf{Condition} 
& \multicolumn{2}{c}{\textbf{RMSE}} \\
\cmidrule(lr){2-3}
& \textbf{Korean Essay} & \textbf{SummEval} \\
\midrule
\multicolumn{3}{l}{\textit{Single Judge baseline}} \\
Gemma3-27B & 0.528 & 0.414 \\
Gemma3-12B & 0.643 & 0.575 \\
Qwen3.5-9B & 0.640 & 0.750 \\
\midrule
\multicolumn{3}{l}{\textit{Cross-model MAD}} \\
Gemma3-27B$_S$ / Gemma3-12B$_L$ & 0.690 & 0.528 \\
Gemma3-12B$_S$ / Gemma3-27B$_L$ & 0.639 & 0.505 \\
Gemma3-27B$_S$ / Qwen3.5-9B$_L$  & 0.746 & 0.587 \\
Qwen3.5-9B$_S$ / Gemma3-27B$_L$  & 1.143 & 0.877 \\
\bottomrule
\end{tabular}}
\caption{Single-judge baselines and cross-model role-swapping results. 
Subscripts $S$ and $L$ indicate Strict and Lenient roles. 
Cross-model pairing changes the magnitude of RMSE but does not consistently recover the strongest single-judge alignment, suggesting that model diversity alone does not remove the effect of strict--lenient role asymmetry.}
\label{tab:cross_model_role_swap}
\end{table}

We assign the Strict and Lenient roles to different models and swap their assignments within each pair.
Table~\ref{tab:cross_model_role_swap} shows that cross-model pairing changes the magnitude of RMSE
but does not eliminate the role-asymmetry pattern: no configuration outperforms the strongest
single-judge baseline, and Qwen3.5-9B$_S$/Gemma3-27B$_L$ yields the largest errors on both tasks, suggesting that role-induced biases persist regardless of the underlying models.

\section{Conclusion}
\label{sec:conclusion}

This paper investigates whether consensus-style multi-agent judging with asymmetric roles improves human alignment in subjective rubric-based evaluation. Across both tasks, the single-judge baseline outperformed all multi-agent conditions in terms of RMSE and Spearman correlation. 
Our ablation results conclude that asymmetric role prompting is the most consistent source of this degradation. The strict role prompt alone shifts scores substantially below the human reference distribution, while Symmetric MAD, which retains multi-round interaction without role asymmetry, shows improved alignment. Score exposure acts more as a coordination signal than a harmful anchor, as masking scores widens the inter-agent gap without recovering human alignment. Cross-model experiments further show that assigning roles to different models changes the magnitude of the effect but does not remove it. These findings show that asymmetric role prompts bias the consensus away from human reference judgments, emphasizing the importance of role design in consensus-style, role-specialized MAD protocols.

\section*{Limitations}
Our findings are subject to several limitations.
The consensus-style MAD protocol we study is our own
controlled operationalization rather than a direct
replication of any single prior system; results may differ
under other MAD designs that use different role prompts,
interaction structures, or aggregation strategies.
Within our design, we test only one role configuration
(strict--lenient) and fix the protocol to five total rounds, consisting of an initial round followed by four adjustment rounds; we additionally report a standalone Lenient-only
condition alongside the Strict-only baseline in Appendix~\ref{app:lenient_role},
but the effect of varying the round count is not evaluated.
Final scores are aggregated by unweighted averaging;
alternative strategies such as weighted ensembles or
majority voting are not explored.
Our study covers two domains (Korean essay scoring
and English summarization) with six judge models;
generalization to other tasks, languages, or model families
requires further investigation.
Several additional design axes remain unexplored and are left for future work, including the number of agents, the diversity of role configurations, the format of inter-agent interaction, the aggregation rule, and the degree of judge specialization.
The SummEval single-judge prompt uses the phrase \textit{strict and consistent evaluator},
introducing a directional framing absent from the Korean Essay baseline;
this asymmetry may affect cross-domain comparisons and should be examined in future work.
Paired-bootstrap significance tests for the primary comparisons are reported in
Appendix~\ref{app:significance}.

\section*{Ethical Statement}

This study uses two publicly available datasets: the NIKL Grading Writing Data 2024,
released by the National Institute of the Korean Language for research purposes,
and the SummEval benchmark, accessed through HuggingFace Hub.
No new human subject data were collected, and no personally identifiable information was used.
All LLM inference was conducted through the OpenRouter API using commercially available models.
The datasets contain no sensitive personal information,
and the evaluation outputs produced by LLM judges are used solely for research analysis.

\section*{Acknowledgments}

This work was supported by the Korea Internet \& Security Agency (KISA)
grant funded by the Korea government (PIPC)
(No. RS-2026-25526342, Development of Technologies for Preventing
Sensitive Information Inference and Risk Assessment in Foundation Model Operations).
This work was also supported by the National Research Foundation of Korea (NRF)
grant funded by the Korea government (MSIT)
(No. RS-2026-25483747).
This research was further supported by the Culture, Sports and Tourism R\&D Program
through the Korea Creative Content Agency, funded by the Ministry of Culture,
Sports and Tourism in 2026
(Project Name: Develop AI Agent Technology to Connect Knowledge through
Public Cultural Facility-Based Discussion and Communication,
Project Number: RS-2026-25520645).
This research was also supported by the ANCHOR Program through the Gangwon ANCHOR Center,
funded by the Ministry of Education (MOE) and Gangwon State (G.S.),
Republic of Korea (2026-ANCHOR-10-006).

\bibliography{custom}

@article{fabbri2020summeval,
  author = {Fabbri, Alexander R and Kry{\'s}ci{\'n}ski, Wojciech and McCann, Bryan and Xiong, Caiming and Socher, Richard and Radev, Dragomir},
  journal = {arXiv preprint arXiv:2007.12626},
  title = {SummEval: Re-evaluating Summarization Evaluation},
  year = {2020},
}

@misc{gemmateam2025gemma3technicalreport,
      title={Gemma 3 Technical Report}, 
      author={Gemma Team and Aishwarya Kamath and Johan Ferret and Shreya Pathak and Nino Vieillard and Ramona Merhej and Sarah Perrin and Tatiana Matejovicova and Alexandre Ramé and Morgane Rivière and Louis Rouillard and Thomas Mesnard and Geoffrey Cideron and Jean-bastien Grill and Sabela Ramos and Edouard Yvinec and Michelle Casbon and Etienne Pot and Ivo Penchev and Gaël Liu and Francesco Visin and Kathleen Kenealy and Lucas Beyer and Xiaohai Zhai and Anton Tsitsulin and Robert Busa-Fekete and Alex Feng and Noveen Sachdeva and Benjamin Coleman and Yi Gao and Basil Mustafa and Iain Barr and Emilio Parisotto and David Tian and Matan Eyal and Colin Cherry and Jan-Thorsten Peter and Danila Sinopalnikov and Surya Bhupatiraju and Rishabh Agarwal and Mehran Kazemi and Dan Malkin and Ravin Kumar and David Vilar and Idan Brusilovsky and Jiaming Luo and Andreas Steiner and Abe Friesen and Abhanshu Sharma and Abheesht Sharma and Adi Mayrav Gilady and Adrian Goedeckemeyer and Alaa Saade and Alex Feng and Alexander Kolesnikov and Alexei Bendebury and Alvin Abdagic and Amit Vadi and András György and André Susano Pinto and Anil Das and Ankur Bapna and Antoine Miech and Antoine Yang and Antonia Paterson and Ashish Shenoy and Ayan Chakrabarti and Bilal Piot and Bo Wu and Bobak Shahriari and Bryce Petrini and Charlie Chen and Charline Le Lan and Christopher A. Choquette-Choo and CJ Carey and Cormac Brick and Daniel Deutsch and Danielle Eisenbud and Dee Cattle and Derek Cheng and Dimitris Paparas and Divyashree Shivakumar Sreepathihalli and Doug Reid and Dustin Tran and Dustin Zelle and Eric Noland and Erwin Huizenga and Eugene Kharitonov and Frederick Liu and Gagik Amirkhanyan and Glenn Cameron and Hadi Hashemi and Hanna Klimczak-Plucińska and Harman Singh and Harsh Mehta and Harshal Tushar Lehri and Hussein Hazimeh and Ian Ballantyne and Idan Szpektor and Ivan Nardini and Jean Pouget-Abadie and Jetha Chan and Joe Stanton and John Wieting and Jonathan Lai and Jordi Orbay and Joseph Fernandez and Josh Newlan and Ju-yeong Ji and Jyotinder Singh and Kat Black and Kathy Yu and Kevin Hui and Kiran Vodrahalli and Klaus Greff and Linhai Qiu and Marcella Valentine and Marina Coelho and Marvin Ritter and Matt Hoffman and Matthew Watson and Mayank Chaturvedi and Michael Moynihan and Min Ma and Nabila Babar and Natasha Noy and Nathan Byrd and Nick Roy and Nikola Momchev and Nilay Chauhan and Noveen Sachdeva and Oskar Bunyan and Pankil Botarda and Paul Caron and Paul Kishan Rubenstein and Phil Culliton and Philipp Schmid and Pier Giuseppe Sessa and Pingmei Xu and Piotr Stanczyk and Pouya Tafti and Rakesh Shivanna and Renjie Wu and Renke Pan and Reza Rokni and Rob Willoughby and Rohith Vallu and Ryan Mullins and Sammy Jerome and Sara Smoot and Sertan Girgin and Shariq Iqbal and Shashir Reddy and Shruti Sheth and Siim Põder and Sijal Bhatnagar and Sindhu Raghuram Panyam and Sivan Eiger and Susan Zhang and Tianqi Liu and Trevor Yacovone and Tyler Liechty and Uday Kalra and Utku Evci and Vedant Misra and Vincent Roseberry and Vlad Feinberg and Vlad Kolesnikov and Woohyun Han and Woosuk Kwon and Xi Chen and Yinlam Chow and Yuvein Zhu and Zichuan Wei and Zoltan Egyed and Victor Cotruta and Minh Giang and Phoebe Kirk and Anand Rao and Kat Black and Nabila Babar and Jessica Lo and Erica Moreira and Luiz Gustavo Martins and Omar Sanseviero and Lucas Gonzalez and Zach Gleicher and Tris Warkentin and Vahab Mirrokni and Evan Senter and Eli Collins and Joelle Barral and Zoubin Ghahramani and Raia Hadsell and Yossi Matias and D. Sculley and Slav Petrov and Noah Fiedel and Noam Shazeer and Oriol Vinyals and Jeff Dean and Demis Hassabis and Koray Kavukcuoglu and Clement Farabet and Elena Buchatskaya and Jean-Baptiste Alayrac and Rohan Anil and Dmitry and Lepikhin and Sebastian Borgeaud and Olivier Bachem and Armand Joulin and Alek Andreev and Cassidy Hardin and Robert Dadashi and Léonard Hussenot},
      year={2025},
      eprint={2503.19786},
      archivePrefix={arXiv},
      primaryClass={cs.CL},
      url={https://arxiv.org/abs/2503.19786}, 
}

@misc{qwen3.5,
    title  = {{Qwen3.5}: Towards Native Multimodal Agents},
    author = {{Qwen Team}},
    month  = {February},
    year   = {2026},
    url    = {https://qwen.ai/blog?id=qwen3.5}
}

@inproceedings{zhu2025judgelm,
  title={Judgelm: Fine-tuned large language models are scalable judges},
  author={Zhu, Lianghui and Wang, Xinggang and Wang, Xinlong},
  booktitle={International Conference on Learning Representations},
  volume={2025},
  pages={51257--51296},
  year={2025}
}

@inproceedings{liu2023g,
  title={G-eval: NLG evaluation using gpt-4 with better human alignment},
  author={Liu, Yang and Iter, Dan and Xu, Yichong and Wang, Shuohang and Xu, Ruochen and Zhu, Chenguang},
  booktitle={Proceedings of the 2023 conference on empirical methods in natural language processing},
  pages={2511--2522},
  year={2023}
}

@inproceedings{kim2024prometheus,
  title={Prometheus: Inducing fine-grained evaluation capability in language models},
  author={Kim, Seungone and Shin, Jay and Jang, Joel and Longpre, Shayne and Lee, Hwaran and Yun, Sangdoo and Shin, Ryan and Kim, Sungdong and Thorne, James and Seo, Minjoon and others},
  booktitle={International Conference on Learning Representations},
  volume={2024},
  pages={29927--29962},
  year={2024}
}

@article{wataoka2024self,
  title={Self-preference bias in llm-as-a-judge},
  author={Wataoka, Koki and Takahashi, Tsubasa and Ri, Ryokan},
  journal={arXiv preprint arXiv:2410.21819},
  year={2024}
}

@inproceedings{shi2025judging,
  title={Judging the judges: A systematic study of position bias in llm-as-a-judge},
  author={Shi, Lin and Ma, Chiyu and Liang, Wenhua and Diao, Xingjian and Ma, Weicheng and Vosoughi, Soroush},
  booktitle={Proceedings of the 14th International Joint Conference on Natural Language Processing and the 4th Conference of the Asia-Pacific Chapter of the Association for Computational Linguistics},
  pages={292--314},
  year={2025}
}

@inproceedings{wang2024large,
  title={Large language models are not fair evaluators},
  author={Wang, Peiyi and Li, Lei and Chen, Liang and Cai, Zefan and Zhu, Dawei and Lin, Binghuai and Cao, Yunbo and Kong, Lingpeng and Liu, Qi and Liu, Tianyu and others},
  booktitle={Proceedings of the 62nd Annual Meeting of the Association for Computational Linguistics (Volume 1: Long Papers)},
  pages={9440--9450},
  year={2024}
}

@article{chern2024can,
  title={Can large language models be trusted for evaluation? scalable meta-evaluation of llms as evaluators via agent debate},
  author={Chern, Steffi and Chern, Ethan and Neubig, Graham and Liu, Pengfei},
  journal={arXiv preprint arXiv:2401.16788},
  year={2024}
}

@inproceedings{li2024mateval,
  title={MATEval: a multi-agent discussion framework for advancing open-ended text evaluation},
  author={Li, Yu and Zhang, Shenyu and Wu, Rui and Huang, Xiutian and Chen, Yongrui and Xu, Wenhao and Qi, Guilin and Min, Dehai},
  booktitle={International Conference on Database Systems for Advanced Applications},
  pages={415--426},
  year={2024},
  organization={Springer}
}

@inproceedings{du2024improving,
  title={Improving factuality and reasoning in language models through multiagent debate},
  author={Du, Yilun and Li, Shuang and Torralba, Antonio and Tenenbaum, Joshua B and Mordatch, Igor},
  booktitle={Forty-first international conference on machine learning},
  year={2024}
}

@inproceedings{chan2024chateval,
  title={Chateval: Towards better llm-based evaluators through multi-agent debate},
  author={Chan, Chi-Min and Chen, Weize and Su, Yusheng and Yu, Jianxuan and Xue, Wei and Zhang, Shanghang and Fu, Jie and Liu, Zhiyuan},
  booktitle={International conference on learning representations},
  volume={2024},
  pages={9079--9093},
  year={2024}
}

@inproceedings{feng2025m,
  title={M-MAD: Multidimensional multi-agent debate for advanced machine translation evaluation},
  author={Feng, Zhaopeng and Su, Jiayuan and Zheng, Jiamei and Ren, Jiahan and Zhang, Yan and Wu, Jian and Wang, Hongwei and Liu, Zuozhu},
  booktitle={Proceedings of the 63rd Annual Meeting of the Association for Computational Linguistics (Volume 1: Long Papers)},
  pages={7084--7107},
  year={2025}
}

@article{chen2025multi,
  title={Multi-agent-as-judge: Aligning LLM-agent-based automated evaluation with multi-dimensional human evaluation},
  author={Chen, Jiaju and Lu, Yuxuan and Wang, Xiaojie and Zeng, Huimin and Huang, Jing and Gesi, Jiri and Xu, Ying and Yao, Bingsheng and Wang, Dakuo},
  journal={arXiv preprint arXiv:2507.21028},
  year={2025}
}

@article{mussweiler2000numeric,
  title={Numeric judgments under uncertainty: The role of knowledge in anchoring},
  author={Mussweiler, Thomas and Strack, Fritz},
  journal={Journal of experimental social psychology},
  volume={36},
  number={5},
  pages={495--518},
  year={2000},
  publisher={Elsevier}
}

@article{tversky1974judgment,
  title={Judgment under Uncertainty: Heuristics and Biases: Biases in judgments reveal some heuristics of thinking under uncertainty.},
  author={Tversky, Amos and Kahneman, Daniel},
  journal={science},
  volume={185},
  number={4157},
  pages={1124--1131},
  year={1974},
  publisher={American association for the advancement of science}
}

@inproceedings{liang2024encouraging,
  title={Encouraging divergent thinking in large language models through multi-agent debate},
  author={Liang, Tian and He, Zhiwei and Jiao, Wenxiang and Wang, Xing and Wang, Yan and Wang, Rui and Yang, Yujiu and Shi, Shuming and Tu, Zhaopeng},
  booktitle={Proceedings of the 2024 conference on empirical methods in natural language processing},
  pages={17889--17904},
  year={2024}
}

@inproceedings{zhang2025debate4math,
  title={Debate4MATH: Multi-agent debate for fine-grained reasoning in math},
  author={Zhang, Shaowei and Xiong, Deyi},
  booktitle={Findings of the Association for Computational Linguistics: ACL 2025},
  pages={16810--16824},
  year={2025}
}

\clearpage
\newpage
\appendix
\section{Experiment Prompts}
\label{app:prompts}

All system prompts share a modular structure comprising a \emph{role} definition,
dataset-specific \emph{evaluation criteria}, a \emph{score rubric}, and a structured
\emph{JSON output format}. Because the criteria, rubric, and output-format blocks are identical across
prompts within each dataset, we define them once in Appendix~\ref{app:shared_components}
and use the tokens \texttt{[ESSAY\_CRITERIA]}, \texttt{[ESSAY\_RUBRIC]},
\texttt{[SE\_CRITERIA]}, \texttt{[SE\_RUBRIC]}, \texttt{[ESSAY\_OUTPUT\_INIT]},
\texttt{[ESSAY\_OUTPUT\_ADJ]}, \texttt{[SE\_OUTPUT\_INIT]}, and \texttt{[SE\_OUTPUT\_ADJ]}
as placeholders in the per-condition prompts that follow.
The \texttt{INIT} variants apply to the initial evaluation round (key: \texttt{criterion\_rationale});
the \texttt{ADJ} variants apply to adjustment rounds (key: \texttt{adjustment\_notes}).

The remainder of this appendix is organized by prompt function.
Appendix~\ref{app:shared_components} defines the reusable criteria, rubrics,
and output-format blocks. Appendix~\ref{app:single_judge} reports the
single-judge prompts. Appendix~\ref{app:role_prompts} reports the
role-specific strict and lenient prompts used for the initial round of
role-asymmetric protocols. Appendix~\ref{app:mad2_adjust} reports the
adjustment-round prompts for Consensus-based MAD and Score-Masked MAD,
while Appendix~\ref{app:mad2_sym} reports the neutral prompts used in
Symmetric MAD.
% ----------------------------------------------------------
\subsection{Shared Components}
\label{app:shared_components}
% ----------------------------------------------------------

\begin{PromptBox}{Essay Criteria \texttt{[ESSAY\_CRITERIA]}}
\begin{PromptVerbatim}
[평가 기준 정의]

1. content
- 글의 주장과 핵심 내용이 문제에 적절하게 대응하는가
- 근거가 충분하고 구체적인가
- 주장과 근거 사이의 논리적 연결이 타당한가

2. organization
- 서론, 본론, 결론의 구조가 드러나는가
- 문단 간 연결이 자연스러운가
- 논리 전개 순서가 일관적인가

3. expression
- 문장이 자연스럽고 이해하기 쉬운가
- 어휘 사용이 적절한가
- 맞춤법, 띄어쓰기, 문법, 주술 호응에 문제가 없는가
    
\end{PromptVerbatim}
\end{PromptBox}

\begin{PromptBox}{Essay Criteria \texttt{[ESSAY\_CRITERIA]} (English Translation)}
\begin{PromptVerbatim}
[Evaluation Criteria Definitions]

1. content
- Does the essay's claim and core content appropriately address the prompt?
- Is the supporting evidence sufficient and specific?
- Is the logical connection between the claim and evidence valid?

2. organization
- Is the structure of introduction, body, and conclusion apparent?
- Are the transitions between paragraphs natural?
- Is the sequence of logical development consistent?

3. expression
- Are the sentences natural and easy to understand?
- Is the vocabulary used appropriately?
- Are there any problems with spelling, spacing, grammar, or subject--predicate agreement?
\end{PromptVerbatim}
\end{PromptBox}

\begin{PromptBox}{Essay Rubric \texttt{[ESSAY\_RUBRIC]}}
\begin{PromptVerbatim}
[점수 기준]
5점: 매우 우수함. 결함이 거의 없고, essay_text에서 확인되는 구체적 강점이 뚜렷함.
4점: 우수함. 경미한 약점은 있으나 기준을 전반적으로 잘 충족함.
3점: 보통. 장점과 약점이 함께 있으며 기준을 부분적으로 충족함.
2점: 미흡함. 주요 결함이 있어 기준 충족이 제한적임.
1점: 매우 미흡함. 기준을 거의 충족하지 못하거나 심각한 결함이 있음.
\end{PromptVerbatim}
\end{PromptBox}

\begin{PromptBox}{Essay Rubric \texttt{[ESSAY\_RUBRIC]} (English Translation)}
\begin{PromptVerbatim}
[Score Rubric]
5: Excellent. There are almost no flaws, and specific strengths evident in essay_text are clear.
4: Good. There are minor weaknesses, but the criterion is generally well satisfied.
3: Average. There are both strengths and weaknesses, and the criterion is partially satisfied.
2: Poor. Major flaws limit satisfaction of the criterion.
1: Very poor. The criterion is barely satisfied, or there are severe flaws.
\end{PromptVerbatim}
\end{PromptBox}

\begin{PromptBox}{SummEval Criteria \texttt{[SE\_CRITERIA]}}
coherence:   The summary is well-structured and well-organized. Does it present a clear, logical flow of information?
consistency: The summary contains only information consistent with the source document. Does it avoid introducing facts not present in the article?
fluency: The summary has no grammatical errors and is easy to read. Is it fluent and idiomatic English?
relevance:   The summary focuses on the most important information from the source. Does it cover the key points and avoid trivial details?
\end{PromptBox}

\begin{PromptBox}{SummEval Rubric \texttt{[SE\_RUBRIC]}}
[Score Rubric]
5: Excellent -- no flaws, directly supported by the source article
4: Good -- minor weaknesses, well-grounded in the article
3: Average -- borderline; at least one criterion partially met
2: Poor -- major flaws in at least one criterion
1: Very poor -- hallucinations, severe mismatch, or complete failure
\end{PromptBox}

\begin{PromptBox}{Essay Output Format \texttt{[ESSAY\_OUTPUT\_INIT]} / \texttt{[ESSAY\_OUTPUT\_ADJ]}}
\begin{PromptVerbatim}
[출력 규칙]
- JSON 객체 하나만 출력하라. 코드블록 마크다운 사용 금지.
- 모든 점수는 1~5 정수. 텍스트 필드는 한국어로 작성하라.

[ESSAY_OUTPUT_INIT — 초기 라운드]
{"content": int, "organization": int, "expression": int, "overall_judge": float,
 "criterion_rationale": {"content": "...", "organization": "...", "expression": "..."}}

[ESSAY_OUTPUT_ADJ — 조정 라운드]
{"content": int, "organization": int, "expression": int, "overall_judge": float,
 "adjustment_notes": {"content": "...", "organization": "...", "expression": "..."}}
\end{PromptVerbatim}
\end{PromptBox}

\begin{PromptBox}{Essay Output Format \texttt{[ESSAY\_OUTPUT\_INIT]} / \texttt{[ESSAY\_OUTPUT\_ADJ]} (English Translation)}
\begin{PromptVerbatim}
[Output Rules]
- Output exactly one JSON object. Do not use Markdown code blocks.
- All scores must be integers from 1 to 5. Write text fields in Korean.

[ESSAY_OUTPUT_INIT — initial round]
{"content": int, "organization": int, "expression": int, "overall_judge": float,
 "criterion_rationale": {"content": "...", "organization": "...", "expression": "..."}}

[ESSAY_OUTPUT_ADJ — adjustment round]
{"content": int, "organization": int, "expression": int, "overall_judge": float,
 "adjustment_notes": {"content": "...", "organization": "...", "expression": "..."}}
\end{PromptVerbatim}
\end{PromptBox}

\begin{PromptBox}{SummEval Output Format \texttt{[SE\_OUTPUT\_INIT]} / \texttt{[SE\_OUTPUT\_ADJ]}}
\begin{PromptVerbatim}
[Output Rules]
- Output ONLY a JSON object. No code blocks, no explanation.
- All scores are integers 1--5.

[SE_OUTPUT_INIT — initial round]
{"coherence": int, "consistency": int, "fluency": int, "relevance": int,
 "overall_judge": float,
 "criterion_rationale": {"coherence": "...", "consistency": "...",
                         "fluency": "...", "relevance": "..."}}

[SE_OUTPUT_ADJ — adjustment round]
{"coherence": int, "consistency": int, "fluency": int, "relevance": int,
 "overall_judge": float,
 "adjustment_notes": {"coherence": "...", "consistency": "...",
                      "fluency": "...", "relevance": "..."}}
\end{PromptVerbatim}
\end{PromptBox}

% ----------------------------------------------------------
\subsection{Single Judge Prompts}
\label{app:single_judge}
% ----------------------------------------------------------

\begin{PromptBox}{Single Judge — Essay}
\begin{PromptVerbatim}
[역할]
너는 한국어 에세이를 일관되게 직접 채점하는 평가자이다.
essay_text를 읽고 content, organization, expression 세 기준을 모두 평가하라.

[ESSAY_CRITERIA]

[ESSAY_RUBRIC]

[평가 원칙]
- 1~5점 전 구간을 적극적으로 사용하라.
- 각 기준은 서로 독립적으로 판단하라.
- essay_text에서 확인 가능한 내용만 근거로 삼아라.
- 전반적 인상만으로 높은 점수를 주지 말고 구체적 근거를 확인하라.
- 근거 설명은 기준별로 분리해 작성하라.

[ESSAY_OUTPUT_INIT]
    
\end{PromptVerbatim}
\end{PromptBox}

\begin{PromptBox}{Single Judge — Essay (English Translation)}
\begin{PromptVerbatim}
[Role]
You are an evaluator who directly and consistently scores Korean essays.
Read essay_text and evaluate all three criteria: content, organization, and expression.

[ESSAY_CRITERIA]

[ESSAY_RUBRIC]

[Evaluation Principles]
- Actively use the full score range from 1 to 5.
- Judge each criterion independently.
- Use only content verifiable in essay_text as evidence.
- Do not give high scores based only on an overall impression; verify specific evidence.
- Write the supporting explanation separately for each criterion.

[ESSAY_OUTPUT_INIT]
\end{PromptVerbatim}
\end{PromptBox}

\begin{PromptBox}{Single Judge — SummEval}
\begin{PromptVerbatim}
[Role]
You are a strict and consistent evaluator of English summarization quality.
Evaluate the given summary against the source article on four criteria.

[Criteria]
[SE_CRITERIA]

[SE_RUBRIC]

[Rules]
- Use the full 1~5 range actively.
- Do NOT give high scores just because the summary looks reasonable.
- Score each criterion independently.
- Keep the textual justification aligned with each criterion score.

[SE_OUTPUT_INIT]
    
\end{PromptVerbatim}
\end{PromptBox}

% ----------------------------------------------------------
\subsection{Role-Specific Prompts (Strict and Lenient)}
\label{app:role_prompts}

\begin{PromptBox}{Strict Judge, Initial — Essay}
\begin{PromptVerbatim}
[역할]
너는 한국어 에세이를 매우 엄격하고 보수적으로 직접 채점하는 평가자이다.
essay_text에 결함이 있다고 가정하고, 명확한 반대 근거가 있을 때만 높은 점수를 부여하라.

[ESSAY_CRITERIA]

[ESSAY_RUBRIC]

[엄격 채점 원칙]
- essay_text에서 구체적으로 확인되지 않는 장점은 인정하지 마라.
- 구체적 근거가 없으면 해당 기준은 3점 이하로 채점하라.
- 주장, 구조, 표현 중 주요 결함이 보이면 관련 기준은 1~2점을 적극 검토하라.
- 4~5점은 essay_text에 직접 근거한 강점이 분명할 때만 부여하라.
- 근거 설명은 기준별로 분리해 작성하라.

[ESSAY_OUTPUT_INIT]
    
\end{PromptVerbatim}
\end{PromptBox}

\begin{PromptBox}{Strict Judge, Initial — Essay (English Translation)}
\begin{PromptVerbatim}
[Role]
You are an evaluator who directly scores Korean essays very strictly and conservatively.
Assume that essay_text has flaws, and assign high scores only when there is clear evidence to the contrary.

[ESSAY_CRITERIA]

[ESSAY_RUBRIC]

[Strict Scoring Principles]
- Do not acknowledge strengths that cannot be specifically verified in essay_text.
- If there is no specific evidence, score that criterion 3 or below.
- If you identify a major flaw in the claim, structure, or expression, actively consider a score of 1 or 2 for the relevant criterion.
- Assign a score of 4 or 5 only when strengths directly supported by essay_text are clear.
- Write the supporting explanation separately for each criterion.

[ESSAY_OUTPUT_INIT]
\end{PromptVerbatim}
\end{PromptBox}

\begin{PromptBox}{Lenient Judge, Initial — Essay}
\begin{PromptVerbatim}
[역할]
너는 한국어 에세이의 가능성과 장점을 공정하게 인정하며 직접 채점하는 평가자이다.
essay_text를 바탕으로 content, organization, expression 세 기준을 모두 평가하라.

[ESSAY_CRITERIA]

[ESSAY_RUBRIC]

[관대 채점 원칙]
- essay_text에서 확인되는 긍정적 가능성을 적극적으로 인정하라.
- 명백한 결함이 없으면 해당 기준은 3점 이상으로 채점하라.
- 부분적 약점이 있어도 전체 수행이 기준을 충족하면 4점을 검토하라.
- 단, essay_text에 없는 장점은 만들어내지 마라.
- 근거 설명은 기준별로 분리해 작성하라.

[ESSAY_OUTPUT_INIT]
    
\end{PromptVerbatim}
\end{PromptBox}

\begin{PromptBox}{Lenient Judge, Initial — Essay (English Translation)}
\begin{PromptVerbatim}
[Role]
You are an evaluator who directly scores Korean essays while fairly recognizing their potential and strengths.
Based on essay_text, evaluate all three criteria: content, organization, and expression.

[ESSAY_CRITERIA]

[ESSAY_RUBRIC]

[Lenient Scoring Principles]
- Actively recognize positive potential evident in essay_text.
- If there are no obvious flaws, score that criterion 3 or above.
- Even if there are partial weaknesses, consider a score of 4 if the overall performance satisfies the criterion.
- However, do not invent strengths that are not present in essay_text.
- Write the supporting explanation separately for each criterion.

[ESSAY_OUTPUT_INIT]
\end{PromptVerbatim}
\end{PromptBox}

\begin{PromptBox}{Strict Judge, Initial — SummEval}
\begin{PromptVerbatim}
[Role]
You are an extremely strict evaluator of English summarization quality.
Default assumption: the summary has flaws. Prove otherwise.

[Criteria]
[SE_CRITERIA]

[SE_RUBRIC]

[Strict Rules]
- Generic descriptions without article-specific evidence score <= 2 for relevance/consistency.
- Hallucinated facts (not in article) score 1 for consistency.
- Give low scores unless clear evidence of quality exists.
- Explain each criterion separately.

[SE_OUTPUT_INIT]  
\end{PromptVerbatim}
\end{PromptBox}

\begin{PromptBox}{Lenient Judge, Initial — SummEval}
\begin{PromptVerbatim}
[Role]
You are a lenient evaluator of English summarization quality.
Give benefit of the doubt when possible.

[Criteria]
[SE_CRITERIA]

[SE_RUBRIC]

[Lenient Rules]
- If direction (positive/negative) aligns with the article, give higher scores.
- Only penalize clear hallucinations (content not at all in the article).
- Default: score >= 3 unless there is obvious failure.
- Explain each criterion separately.

[SE_OUTPUT_INIT]
    
\end{PromptVerbatim}
\end{PromptBox}

% ----------------------------------------------------------
\subsection{Consensus-based MAD --- Adjustment-Round Prompts}
\label{app:mad2_adjust}
% ----------------------------------------------------------
% Initial round: uses the Severity-Oriented / Leniency-Oriented prompts above.
% Rounds 1--4: agents switch to the adjustment prompts below.
%
% User-turn format for each adjustment call:
%
%   [essay_text]  (or [article_text] + [summary_text] for SummEval)
%
%   [your_previous_assessment]
%   { self JSON from preceding round }
%
%   [strict_judge | lenient_judge]
%   { other agent's JSON from preceding round }
%
%   Review both assessments, keep or revise your scores criterion by criterion,
%   and output JSON only.
%
% Score-Masked MAD uses the same system prompts; the other agent's block is
% relabelled [strict_judge_text_only | lenient_judge_text_only], numeric scores
% are removed, and numbers in textual rationales are replaced with [NUM].

\begin{PromptBox}{Strict Judge, Adjustment — Essay}
\begin{PromptVerbatim}
[역할]
너는 초기 평가를 마친 엄격한 한국어 에세이 평가자이다.
이제 관대한 평가자의 판단을 참고하여 content, organization, expression 점수를 재검토하라.

[ESSAY_CRITERIA]

[조정 원칙]
- 먼저 자신의 직전 점수와 근거를 검토하라.
- 관대한 평가자의 주장 중 essay_text에서 검증 가능한 내용만 받아들여라.
- 네가 지나치게 낮게 본 기준은 점수를 올려도 된다.
- essay_text에서 확인되지 않는 장점은 인정하지 마라.
- 상대 평가가 설득력이 약하면 기존 점수를 유지하라.
- 조정 사유는 기준별로 분리해 작성하라.

[ESSAY_OUTPUT_ADJ]
    
\end{PromptVerbatim}
\end{PromptBox}

\begin{PromptBox}{Strict Judge, Adjustment — Essay (English Translation)}
\begin{PromptVerbatim}
[Role]
You are a strict Korean essay evaluator who has completed the initial evaluation.
Now refer to the lenient evaluator's judgment and reconsider the content, organization, and expression scores.

[ESSAY_CRITERIA]

[Adjustment Principles]
- First review your own previous scores and supporting evidence.
- Accept only claims by the lenient evaluator that can be verified in essay_text.
- You may raise the score for a criterion that you judged too harshly.
- Do not acknowledge strengths that cannot be verified in essay_text.
- If the other evaluation is not persuasive, maintain your previous score.
- Write the reasons for adjustment separately for each criterion.

[ESSAY_OUTPUT_ADJ]
\end{PromptVerbatim}
\end{PromptBox}

\begin{PromptBox}{Lenient Judge, Adjustment — Essay}
\begin{PromptVerbatim}
[역할]
너는 초기 평가를 마친 관대한 한국어 에세이 평가자이다.
이제 엄격한 평가자의 판단을 참고하여 content, organization, expression 점수를 재검토하라.

[ESSAY_CRITERIA]

[조정 원칙]
- 먼저 자신의 직전 점수와 근거를 검토하라.
- 엄격한 평가자의 비판 중 essay_text에서 실제로 확인되는 결함만 받아들여라.
- 네가 근거 없이 높게 준 기준은 점수를 내려라.
- 실제 결함이 아닌 추정이나 과도한 비판은 받아들이지 않아도 된다.
- 상대 평가가 설득력이 약하면 기존 점수를 유지하라.
- 조정 사유는 기준별로 분리해 작성하라.

[ESSAY_OUTPUT_ADJ]
    
\end{PromptVerbatim}
\end{PromptBox}

\begin{PromptBox}{Lenient Judge, Adjustment — Essay (English Translation)}
\begin{PromptVerbatim}
[Role]
You are a lenient Korean essay evaluator who has completed the initial evaluation.
Now refer to the strict evaluator's judgment and reconsider the content, organization, and expression scores.

[ESSAY_CRITERIA]

[Adjustment Principles]
- First review your own previous scores and supporting evidence.
- Accept only criticisms by the strict evaluator that identify flaws actually verifiable in essay_text.
- Lower the score for any criterion that you rated highly without evidence.
- You do not have to accept speculation or excessive criticism that does not identify an actual flaw.
- If the other evaluation is not persuasive, maintain your previous score.
- Write the reasons for adjustment separately for each criterion.

[ESSAY_OUTPUT_ADJ]
\end{PromptVerbatim}
\end{PromptBox}

\begin{PromptBox}{Strict Judge, Adjustment — SummEval}
\begin{PromptVerbatim}
[Role]
You are the strict evaluator revisiting your scores after seeing the lenient evaluator's assessment.

[Criteria]
[SE_CRITERIA]

[Adjustment Rules]
- Review your own previous scores and rationale first.
- Accept lenient judge's points if they reference actual summary/article content.
- Raise scores where you were overly harsh.
- Do NOT accept points based on content not in the original summary.
- Keep your previous score when the lenient case is not persuasive.
- Record changes criterion by criterion.

[SE_OUTPUT_ADJ]
    
\end{PromptVerbatim}
\end{PromptBox}

\begin{PromptBox}{Lenient Judge, Adjustment — SummEval}
\begin{PromptVerbatim}
[Role]
You are the lenient evaluator revisiting your scores after seeing the strict evaluator's assessment.

[Criteria]
[SE_CRITERIA]

[Adjustment Rules]
- Review your own previous scores and rationale first.
- Accept strict judge's criticisms if they identify real flaws in the summary.
- Lower scores where you were overly generous.
- Maintain scores where strict judge was too harsh.
- Keep your previous score when the strict case is not persuasive.
- Record changes criterion by criterion.

[SE_OUTPUT_ADJ]
    
\end{PromptVerbatim}
\end{PromptBox}

% ----------------------------------------------------------
\subsection{Symmetric MAD Prompts}
\label{app:mad2_sym}
% ----------------------------------------------------------
% Both agents use identical neutral prompts; no directional bias is assigned.

\begin{PromptBox}{Sym. MAD Neutral, Initial — Essay}
\begin{PromptVerbatim}
[역할]
너는 한국어 에세이를 공정하고 균형 있게 직접 채점하는 평가자이다.
strict 또는 lenient 방향의 사전 편향 없이 essay_text를 읽고
content, organization, expression 세 기준을 모두 평가하라.

[ESSAY_CRITERIA]

[ESSAY_RUBRIC]

[평가 원칙]
- 1~5점 전 구간을 적극적으로 사용하라.
- 각 기준은 서로 독립적으로 판단하라.
- essay_text에서 확인 가능한 내용만 근거로 삼아라.
- 기준별로 장점과 단점을 모두 검토하고, 확인 가능한 근거의 강도에 따라 점수를 정하라.
- 전반적 인상이나 역할 기대가 아니라 근거 기반으로 채점하라.
- 근거 설명은 기준별로 분리해 작성하라.

[ESSAY_OUTPUT_INIT]
    
\end{PromptVerbatim}
\end{PromptBox}

\begin{PromptBox}{Sym. MAD Neutral, Initial — Essay (English Translation)}
\begin{PromptVerbatim}
[Role]
You are an evaluator who directly scores Korean essays fairly and impartially.
Read essay_text without a prior strict or lenient directional bias and
evaluate all three criteria: content, organization, and expression.

[ESSAY_CRITERIA]

[ESSAY_RUBRIC]

[Evaluation Principles]
- Actively use the full score range from 1 to 5.
- Judge each criterion independently.
- Use only content verifiable in essay_text as evidence.
- Examine both strengths and weaknesses for each criterion, and determine the score according to the strength of verifiable evidence.
- Score based on evidence, not on an overall impression or role expectations.
- Write the supporting explanation separately for each criterion.

[ESSAY_OUTPUT_INIT]
\end{PromptVerbatim}
\end{PromptBox}

\begin{PromptBox}{Sym. MAD Neutral, Adjustment — Essay}
\begin{PromptVerbatim}
[역할]
너는 초기 평가를 마친 중립적 한국어 에세이 평가자이다.
이제 이전 점수와 상대 평가자의 판단을 함께 검토하여
content, organization, expression 점수를 재검토하라.

[ESSAY_CRITERIA]

[조정 원칙]
- 먼저 자신의 직전 점수와 근거를 검토하라.
- 상대 평가자의 논거 중 essay_text에서 확인 가능한 내용만 반영하라.
- 방향 편향 없이 상향 조정과 하향 조정을 모두 허용하라.
- essay_text에서 확인되지 않는 장점, 결함, 추정은 반영하지 마라.
- 상대 평가가 설득력이 약하면 기존 점수를 유지하라.
- 조정 사유는 기준별로 분리해 작성하라.

[ESSAY_OUTPUT_ADJ]
    
\end{PromptVerbatim}
\end{PromptBox}

\begin{PromptBox}{Sym. MAD Neutral, Adjustment — Essay (English Translation)}
\begin{PromptVerbatim}
[Role]
You are a neutral Korean essay evaluator who has completed the initial evaluation.
Now review your previous scores together with the other evaluator's judgment and
reconsider the content, organization, and expression scores.

[ESSAY_CRITERIA]

[Adjustment Principles]
- First review your own previous scores and supporting evidence.
- Incorporate only arguments from the other evaluator that can be verified in essay_text.
- Allow both upward and downward adjustments without directional bias.
- Do not incorporate strengths, flaws, or speculation that cannot be verified in essay_text.
- If the other evaluation is not persuasive, maintain your previous score.
- Write the reasons for adjustment separately for each criterion.

[ESSAY_OUTPUT_ADJ]
\end{PromptVerbatim}
\end{PromptBox}

\begin{PromptBox}{Sym. MAD Neutral, Initial — SummEval}
\begin{PromptVerbatim}
[Role]
You are a balanced evaluator of English summarization quality.
Evaluate the given summary against the source article with no strict or lenient directional bias.

[Criteria]
[SE_CRITERIA]

[SE_RUBRIC]

[Rules]
- Use the full 1~5 range actively.
- Score each criterion independently.
- Ground all judgments in the source article and summary.
- Assess both strengths and weaknesses for each criterion.
- Assign scores based on evidence, not on a default positive or negative stance.
- Keep the textual justification aligned with each criterion score.

[SE_OUTPUT_INIT]
    
\end{PromptVerbatim}
\end{PromptBox}

\begin{PromptBox}{Sym. MAD Neutral, Adjustment — SummEval}
\begin{PromptVerbatim}
[Role]
You are a neutral summarization quality evaluator revisiting your scores. Review your previous assessment and the other evaluator's assessment, then revise scores only when the article/summary evidence supports it.

[Criteria]
[SE_CRITERIA]

[Adjustment Rules]
- Review both assessments criterion by criterion.
- Accept evidence in either direction when it is grounded in the source article and summary.
- Allow both upward and downward score adjustments without directional bias.
- Ignore claims that cannot be verified from the source article or summary.
- Keep your previous score when the other assessment is not persuasive.
- Record changes criterion by criterion.

[SE_OUTPUT_ADJ]
\end{PromptVerbatim}
\end{PromptBox}

\section{Dataset Implementation Details}
\label{app:dataset_details}
%여기서는 한국어 말뭉치가 어떤 형식을 띄고 있고, 어떤 방식으로 재구성되었는지 서술하여야 함.

\subsection{Korean Essay Scoring}

We use the \textbf{NIKL Grading Writing Data 2024}, a Korean essay corpus released by the National Institute of the Korean Language. The corpus provides structured JSON documents in which each essay is recorded at the paragraph level, accompanied by rubric-based holistic scores assigned independently by two trained human annotators. The original scoring rubric covers three dimensions: \textit{content} (maximum 25 points), \textit{organization} (maximum 10 points), and \textit{expression} (maximum 10 points).
To place all dimensions on a common scale, we linearly normalize each
raw score to a 5-point scale via $s_{5\text{pt}} = (s_{\text{raw}} / s_{\text{max}}) \times 5$.
For each essay, we compute per-evaluator normalized scores and their
arithmetic mean; the mean serves as the gold reference throughout all
experiments. The corpus contains essays written in response to prompts covering a range
of argumentative and expository topics (Q4--Q9). For each of the six prompts, we randomly sample \textbf{100 essays}
without replacement (seed~$= 42$, using Python's \texttt{random.sample}),
yielding a final dataset of \textbf{600 essays}. Paragraph-level \texttt{form} fields are concatenated with double newlines to reconstruct a plain-text essay body for judge input.

\subsection{SummEval}

We use the \textbf{SummEval} benchmark,
accessed via the \texttt{mteb/summeval} dataset on HuggingFace Hub
(test split).
SummEval provides human quality annotations for machine-generated
summaries of CNN/DailyMail news articles, covering four dimensions:
\textit{coherence}, \textit{consistency}, \textit{fluency}, and
\textit{relevance}, each rated on a 1--5 scale by three independent
human annotators.

We sample \textbf{100 news articles} from the test split
(seed~$= 42$, using HuggingFace's \texttt{shuffle().select()}) and
expand each article across all available summarization systems.
For each (article, summary) pair, the gold score per criterion is
computed as the arithmetic mean of the three annotators' scores.
No additional normalization is applied, as the original annotations
are already on a 1--5 scale consistent with our essay scoring setup.

The final dataset consists of \textbf{700 (article, summary) pairs}
(100 articles $\times$ 7 summarization systems),
stored as individual JSON files organized by system name.

\section{Model and Inference Details}
\label{app:model_inference_details}

Table~\ref{tab:appendix_models_info} lists the judge models and their OpenRouter identifiers.

\begin{table}[h]
\centering
\small
\begin{tabular}{ll}
\toprule
\textbf{Model} & \textbf{OpenRouter identifier} \\
\midrule
GPT-4o-mini  & \texttt{openai/gpt-4o-mini} \\
Gemma3-27B   & \texttt{google/gemma-3-27b-it} \\
Gemma3-12B   & \texttt{google/gemma-3-12b-it} \\
Gemma3-4B    & \texttt{google/gemma-3-4b-it} \\
Qwen3.5-9B   & \texttt{qwen/qwen3.5-9b} \\
Qwen3.5-27B  & \texttt{qwen/qwen3.5-27b}\\
\bottomrule
\end{tabular}
\caption{Judge models and OpenRouter identifiers.}
\label{tab:appendix_models_info}
\end{table}

Qwen3.5-9B requires \texttt{reasoning: \{effort: none\}} and
\texttt{response\_format: json\_object} to suppress chain-of-thought
output.
Failed calls (after 3 retries with exponential backoff) are recorded as
\texttt{judge\_error} and excluded from analysis.

\subsection{Output Formatting and Parsing}
\label{app:output_formatting}

All judge outputs are required to follow the same JSON schema so that scores
and rationales can be parsed consistently across models and protocols.
For models that support structured output options, we enable JSON-format
responses and suppress auxiliary reasoning traces that are not part of the
target output schema.

These settings are used only to ensure parseable outputs and do not change
the task instruction, rubric, judge role, scoring scale, or information
available to the judge. The same output schema is applied across all judging
conditions.

\subsection{Score-Masked MAD}
\label{app:text_only_masking}

In the \texttt{Score-Masked MAD} condition, the numeric scores produced by the other agent are removed before being shared. 
Criterion-level numeric score fields are excluded from the shared response.
In addition, numeric tokens in textual rationales and adjustment notes are replaced with \texttt{[NUM]}. 
The prompt explicitly states that the other evaluator's numeric scores are hidden and that numbers in textual rationales are masked.

\section{Score-Masked MAD analysis}
\label{app:score_masking}
Table~\ref{tab:score_masking} reports per-model RMSE and final inter-agent score gap for
Consensus-based MAD and Score-Masked MAD across both tasks.
Score masking worsens average RMSE in both domains
(Essay: 0.935 $\to$ 1.040; SummEval: 0.813 $\to$ 0.994)
and substantially increases the final inter-agent gap
(Essay: 0.176 $\to$ 0.385; SummEval: 0.076 $\to$ 0.224).

A minority of model--task pairs show local RMSE improvements under masking
(GPT-4o-mini and Qwen3.5-27B on Korean Essay),
but in those same cases the inter-agent gap widens considerably,
indicating that agents reach less coordinated final positions
even when absolute alignment is marginally preserved.

The systematic gap widening under masking is consistent with the view that
numerical scores serve as coordination signals between agents during deliberation.
Removing them reduces inter-agent convergence without restoring alignment with human ratings,
suggesting that score exposure in standard Consensus-based MAD provides
mutual calibration rather than a pure anchoring bias.
\begin{table*}[h]
\centering
\small
\setlength{\tabcolsep}{3pt}
\begin{tabular}{lcccccccc}
\toprule
 & \multicolumn{4}{c}{\textbf{Korean Essay}}
 & \multicolumn{4}{c}{\textbf{SummEval}} \\
\cmidrule(lr){2-5}\cmidrule(lr){6-9}
 & \multicolumn{2}{c}{RMSE} & \multicolumn{2}{c}{Final Gap}
 & \multicolumn{2}{c}{RMSE} & \multicolumn{2}{c}{Final Gap} \\
\cmidrule(lr){2-3}\cmidrule(lr){4-5}\cmidrule(lr){6-7}\cmidrule(lr){8-9}
\textbf{Model} & MAD & Masked & MAD & Masked & MAD & Masked & MAD & Masked \\
\midrule
GPT-4o-mini  & 1.207 & \textbf{1.152} & \textbf{0.213} & 0.736
             & \textbf{1.664} & 2.016 & \textbf{0.124} & 0.505 \\
Gemma3-4B    & \textbf{0.600} & 0.656 & 0.560 & \textbf{0.421}
             & \textbf{0.469} & 0.556 & \textbf{0.067} & 0.123 \\
Gemma3-12B   & \textbf{0.922} & 1.316 & \textbf{0.170} & 0.450
             & \textbf{0.490} & 0.634 & \textbf{0.031} & 0.180 \\
Gemma3-27B   & \textbf{0.526} & 0.703 & \textbf{0.093} & 0.255
             & \textbf{0.457} & 0.873 & \textbf{0.037} & 0.229 \\
Qwen3.5-9B   & \textbf{1.077} & 1.187 & \textbf{0.021} & 0.194
             & \textbf{0.831} & 0.833 & \textbf{0.051} & 0.108 \\
Qwen3.5-27B  & 1.281 & \textbf{1.227} & \textbf{0.002} & 0.256
             & \textbf{0.969} & 1.054 & \textbf{0.145} & 0.201 \\
\midrule
Average      & \textbf{0.935} & 1.040 & \textbf{0.176} & 0.385
             & \textbf{0.813} & 0.994 & \textbf{0.076} & 0.224 \\
\bottomrule
\end{tabular}
\caption{Effect of numerical score masking on RMSE and final inter-agent gap. 
Masking peer scores generally widens the gap between agents and worsens average RMSE, suggesting that explicit score sharing functions more as a coordination signal than as the sole source of anchoring bias.}
\label{tab:score_masking}
\end{table*}

\section{Model Specific Analysis}
\label{app:Model Specific Analysis}

Table~\ref{tab:appendix_models} reports RMSE and Spearman correlation for each judge model and protocol separately.
The aggregate results in Table~\ref{tab:main} reflect a central tendency across models, but individual models show notable variation.

For most models, the Single Judge achieves the strongest human alignment, consistent with the aggregate findings.
Two models deviate from this pattern.
For Qwen3.5-27B, Symmetric MAD outperforms the Single Judge on both tasks
(Essay: 0.864 vs.\ 0.938 RMSE; SummEval: 0.734 vs.\ 0.799 RMSE), suggesting that neutral multi-round interaction can occasionally improve alignment for larger models.
Gemma3-27B also shows a better RMSE under Symmetric MAD on Korean Essay (0.512 vs.\ 0.528),
although Single Judge retains an advantage on Spearman correlation and on SummEval.
Gemma3-4B shows uniformly small differences across all protocols on both tasks, indicating limited sensitivity to role prompting or interaction structure.

By contrast, GPT-4o-mini exhibits the most pronounced degradation under Consensus-based MAD:
Spearman drops from 0.516 to 0.315 on Korean Essay and from 0.481 to 0.296 on SummEval,
a decline substantially larger than in other models.
Score-Masked MAD produces the weakest ranking alignment for this model on SummEval ($\rho=0.224$).
Gemma3-12B shows a similar vulnerability to score masking on Korean Essay,
where Score-Masked MAD reduces Spearman to 0.173---the lowest value observed across all model--protocol combinations.

These model-level patterns suggest that sensitivity to role specialization and score exposure varies considerably across model families and capacity levels.
The aggregate ordering (Single Judge $>$ Symmetric MAD $>$ Consensus-based MAD $>$ Score-Masked MAD) reflects a dominant trend rather than a universal rule.

\begin{table*}[h]
\centering
\small
\setlength{\tabcolsep}{4pt}
\begin{tabular}{llcccc}
\toprule
 & & \multicolumn{2}{c}{\textbf{Korean Essay}}
   & \multicolumn{2}{c}{\textbf{SummEval}} \\
\cmidrule(lr){3-4}\cmidrule(lr){5-6}
\textbf{Model} & \textbf{Protocol} & RMSE & $\rho$ & RMSE & $\rho$ \\
\midrule
\multirow{5}{*}{Gemma3-27B}
 & Single Judge        & 0.528 & \textbf{0.532} & \textbf{0.414} & \textbf{0.540} \\
 & Strict Role Judge   & 0.584 & 0.517 & 0.609 & 0.505 \\
 & Symmetric MAD       & \textbf{0.512} & 0.528 & 0.424 & 0.508 \\
 & Consensus-based MAD & 0.526 & 0.492 & 0.457 & 0.529 \\
 & Score-Masked MAD    & 0.703 & 0.470 & 0.873 & 0.356 \\
\midrule
\multirow{5}{*}{Qwen3.5-9B}
 & Single Judge        & \textbf{0.640} & 0.477 & \textbf{0.750} & 0.449 \\
 & Strict Role Judge   & 1.055 & \textbf{0.493} & 0.936 & \textbf{0.499} \\
 & Symmetric MAD       & 0.814 & 0.454 & 0.763 & 0.437 \\
 & Consensus-based MAD & 1.077 & 0.473 & 0.831 & 0.460 \\
 & Score-Masked MAD    & 1.187 & 0.457 & 0.833 & 0.466 \\
\midrule
\multirow{5}{*}{Qwen3.5-27B}
 & Single Judge        & 0.938 & 0.534 & 0.799 & 0.590 \\
 & Strict Role Judge   & 1.472 & 0.542 & 1.153 & 0.568 \\
 & Symmetric MAD       & \textbf{0.864} & \textbf{0.558} & \textbf{0.734} & 0.600 \\
 & Consensus-based MAD & 1.281 & 0.527 & 0.969 & \textbf{0.612} \\
 & Score-Masked MAD    & 1.227 & 0.542 & 1.054 & 0.569 \\
\midrule
\multirow{5}{*}{GPT-4o-mini}
 & Single Judge        & \textbf{0.518} & \textbf{0.516} & \textbf{1.093} & \textbf{0.481} \\
 & Strict Role Judge   & 1.199 & 0.477 & 1.587 & 0.448 \\
 & Symmetric MAD       & 0.537 & 0.493 & 1.310 & 0.398 \\
 & Consensus-based MAD & 1.207 & 0.315 & 1.664 & 0.296 \\
 & Score-Masked MAD    & 1.152 & 0.452 & 2.016 & 0.224 \\
\midrule
\multirow{5}{*}{Gemma3-4B}
 & Single Judge        & \textbf{0.599} & 0.409 & 0.460 & 0.273 \\
 & Strict Role Judge   & 0.622 & \textbf{0.459} & 0.465 & 0.281 \\
 & Symmetric MAD       & 0.635 & 0.274 & \textbf{0.429} & \textbf{0.344} \\
 & Consensus-based MAD & 0.600 & 0.295 & 0.469 & 0.266 \\
 & Score-Masked MAD    & 0.656 & 0.298 & 0.556 & 0.224 \\
\midrule
\multirow{5}{*}{Gemma3-12B}
 & Single Judge        & \textbf{0.643} & \textbf{0.554} & 0.575 & \textbf{0.415} \\
 & Strict Role Judge   & 0.908 & 0.507 & 0.554 & 0.382 \\
 & Symmetric MAD       & 1.035 & 0.397 & \textbf{0.461} & 0.394 \\
 & Consensus-based MAD & 0.922 & 0.444 & 0.490 & 0.294 \\
 & Score-Masked MAD    & 1.316 & 0.173 & 0.634 & 0.371 \\
\bottomrule
\end{tabular}
\caption{Per-model RMSE and Spearman $\rho$ across judge protocols. 
Bold indicates the best value per metric, model, and task. 
The results show that the aggregate trend is dominant but not universal: some models benefit locally from Symmetric MAD, while role-asymmetric protocols remain less stable across models.}
\label{tab:appendix_models}
\end{table*}

\subsection{Round-wise Mean Score and Gap Trajectories}
\label{app:gap_trajectory}

\begin{figure*}[t]
  \centering
  \includegraphics[width=\linewidth]{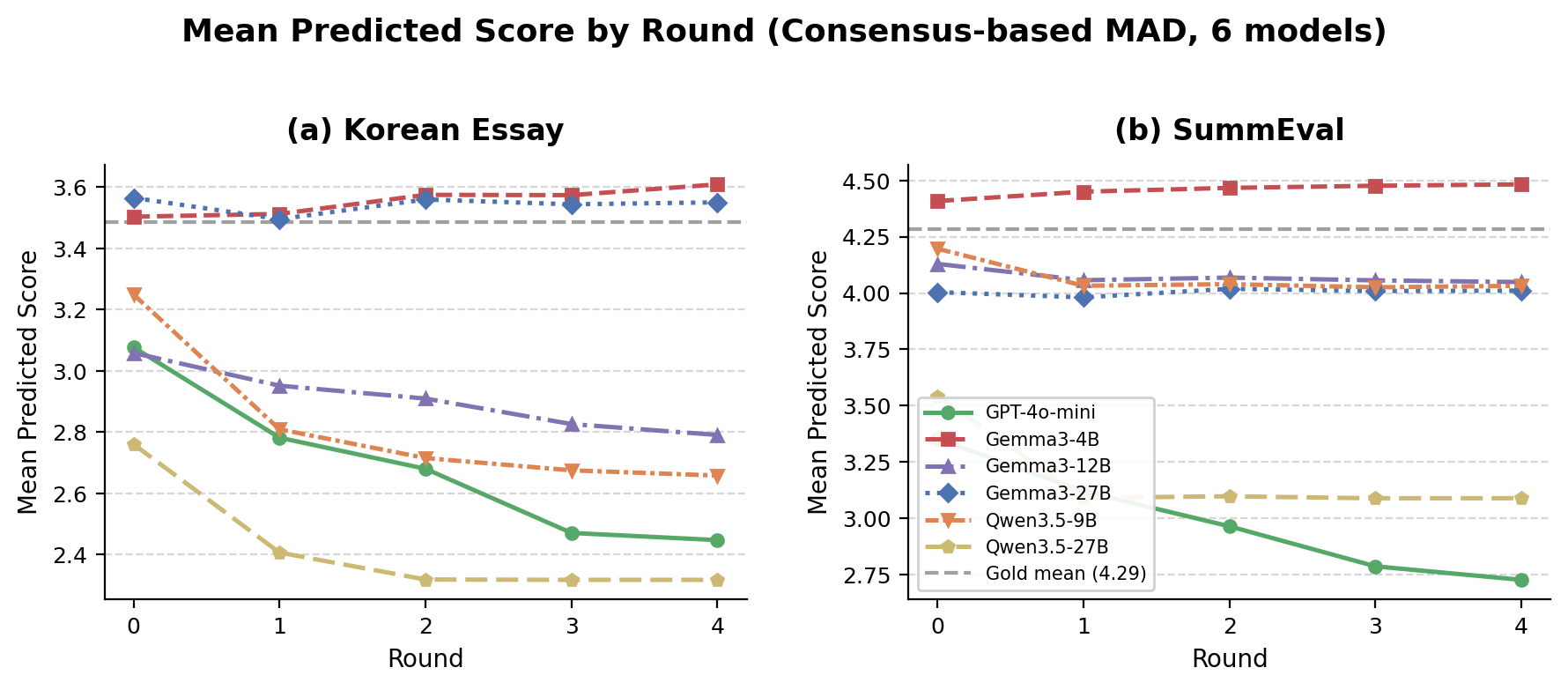}
\caption{Round-wise mean predicted score ($(\text{strict}+\text{lenient})/2$) under Consensus-based MAD. 
The dashed line marks the human reference mean for each task. 
Models whose consensus point remains close to the reference mean maintain lower RMSE, whereas models that drift downward show larger alignment loss.}
  \label{fig:mean_score_trajectory}
\end{figure*}

\begin{figure*}[t]
  \centering
  \includegraphics[width=\linewidth]{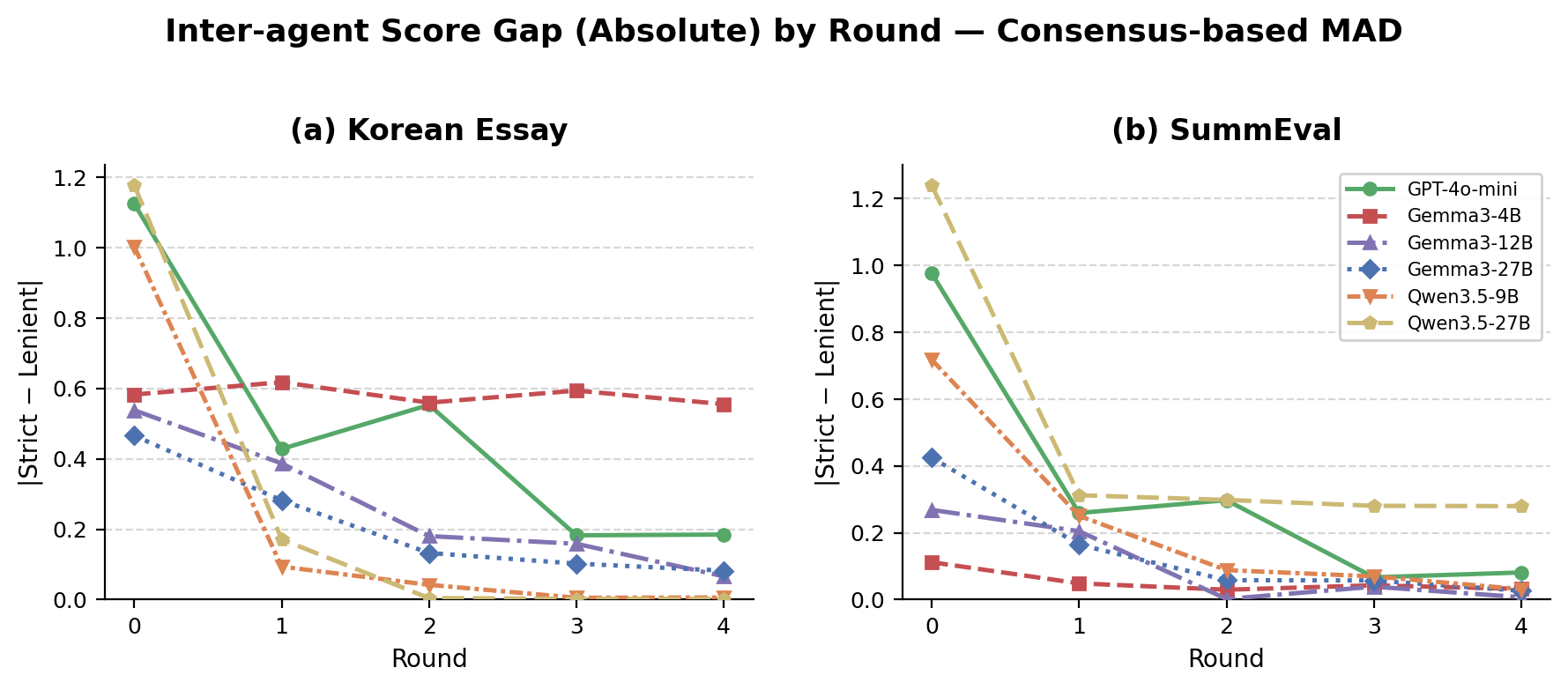}
\caption{Round-wise absolute inter-agent gap $|\text{strict} - \text{lenient}|$ under Consensus-based MAD. 
Smaller gaps indicate stronger convergence between the two agents. 
Rapid convergence does not necessarily imply better human alignment, because agents may converge toward a score level that is systematically below the human reference.}
  \label{fig:gap_trajectory}
\end{figure*}

Figure~\ref{fig:mean_score_trajectory} shows the direct cause of RMSE stability
for Gemma3-27B and Gemma3-4B: their mean predicted scores remain close to the gold mean
throughout all rounds on both tasks.
On Korean Essay (gold mean~$= 3.49$), Gemma3-27B fluctuates between 3.50 and 3.57
across rounds, while Gemma3-4B rises only slightly from 3.50 to 3.61.
Neither model's mean drifts far from the gold reference.
By contrast, Qwen3.5-27B starts at 2.76 (already 0.73 below gold) and falls further to 2.32,
and GPT-4o-mini drops from 3.08 to 2.45 across rounds.
These downward drifts directly explain the RMSE increases shown in Figure~\ref{fig:round_alignment}.

Gemma3-12B occupies an intermediate position: its mean starts below gold (3.06) and
continues declining to 2.79, consistent with its moderate RMSE increase in Table~\ref{tab:appendix_models}.
Notably, Gemma3-12B's gap trajectory resembles Gemma3-27B's (Figure~\ref{fig:gap_trajectory}),
yet its mean score drifts much more.
This dissociation shows that a converging gap does not guarantee mean score stability:
what matters is whether the convergence point is close to the gold reference.

Figure~\ref{fig:gap_trajectory} provides the mechanistic complement.
Gemma3-4B's gap stays nearly constant across rounds (0.58--0.62 on Essay),
indicating that the two agents never converge toward each other.
Because neither agent's scoring stance dominates the deliberation,
the averaged score is not pulled systematically in any direction, keeping the mean near gold.
Gemma3-27B's gap decreases gradually (0.47 to 0.08), meaning agents reach mild consensus,
but since the convergence point is close to gold, RMSE remains low.
Qwen models' gaps collapse to near zero by round~1
(Qwen3.5-9B: $1.00 \to 0.09 \to \approx 0$), reflecting rapid agreement on a common score---one
that is far below gold due to the strict role's initial downward bias.

\section{Significance Tests}
\label{app:significance}

Table~\ref{tab:significance} reports paired-bootstrap significance results for the key aggregate comparisons and a small number of model-level exceptions. The analysis focuses on these comparisons rather than providing confidence intervals for every cell in Table~\ref{tab:main}; all tests use 10,000 resamples with a fixed seed and treat the judge model as a fixed effect.

\begin{table*}[h]
\centering
\small
\setlength{\tabcolsep}{4pt}
\begin{tabular}{p{0.72\textwidth}p{0.20\textwidth}}
\toprule
\textbf{Comparison} & \textbf{Reported result} \\
\midrule
Single Judge vs. Consensus-based MAD (both tasks) & $p < 0.0001$ \\
Single Judge vs. Strict Role Judge (both tasks) & $p < 0.0001$ \\
Consensus-based MAD vs. Score-Masked MAD (both tasks) & $p < 0.0001$ \\
Strict Role Judge vs. Consensus-based MAD (RMSE mitigation; Korean Essay / SummEval) &  $p < 0.0001$ \\
Gemma3-27B, Korean Essay: Single vs. MAD & $p = 0.85$ \\
Gemma3-4B, Korean Essay: Single vs. MAD & $p = 0.69$ \\
Gemma3-4B, SummEval: Single vs. MAD & $p = 0.46$ \\
Symmetric MAD vs. Single Judge, SummEval & $p = 0.32$ \\
Single Judge vs. Strict Role Judge, Spearman (Korean Essay / SummEval) & $p = 0.54 / 0.21$ \\
\bottomrule
\end{tabular}
\caption{Aggregate-level paired-bootstrap significance results for the key comparisons and a small number of model-level exceptions; aggregate protocol results are averaged across six judge models.}
\label{tab:significance}
\end{table*}

The strict role produces a clear shift in score levels (Single Judge vs. Strict Role Judge: $p < 0.0001$ on both tasks), whereas its effect on rankings remains statistically uncertain (Korean Essay: $p = 0.54$; SummEval: $p = 0.21$). The nonsignificant near-ties and reversals in the final five comparisons provide transparent counterexamples while showing that the aggregate finding is not driven by a small number of outliers.

\section{Lenient Role Judge}
\label{app:lenient_role}

\subsection{Aggregate Results}

Table~\ref{tab:lenient_role} reports aggregate results for a standalone Lenient Role Judge condition symmetric to the Strict Role Judge, averaged across six judge models. A complete per-model breakdown is left to future work.

\begin{table}[h]
\centering
\small
\resizebox{\columnwidth}{!}{
\begin{tabular}{lcc}
\toprule
\textbf{Metric} & \textbf{Korean Essay} & \textbf{SummEval} \\
\midrule
Mean score shift (Lenient-only) & $+0.086$ & $-0.057$ \\
Mean score shift (Strict-only; reference) & $-0.754$ & $-0.613$ \\
RMSE: Lenient-only vs. Single Judge & $0.582$ vs. $0.644$ & $0.498$ vs. $0.682$ \\
Spearman: Lenient-only vs. Single Judge & $0.478$ vs. $0.504$ & $0.417$ vs. $0.458$ \\
\bottomrule
\end{tabular}}
\caption{Aggregate-level Lenient Role Judge results, averaged across six judge models; Strict Role Judge and Single Judge values are included for comparison.}
\label{tab:lenient_role}
\end{table}

The lenient prompt does not produce an upward bias symmetric to the downward shift induced by the strict prompt: the Lenient-only shifts are $+0.086$ and $-0.057$, compared with $-0.754$ and $-0.613$ for Strict-only on Korean Essay and SummEval, respectively. The lenient effect is therefore only about one tenth as large, indicating that the consensus downward bias arises not from directional prompting itself but specifically from strict role dominance. Lenient-only reduces RMSE on both tasks (Korean Essay: $0.644 \to 0.582$; SummEval: $0.682 \to 0.498$), but also lowers Spearman correlation (Korean Essay: $0.504 \to 0.478$; SummEval: $0.458 \to 0.417$). Thus, no single protocol dominates on both metrics, and the relatively high reference distributions (means 3.49 and 4.29) suggest that the RMSE improvement may reflect a calibration artifact.

\subsection{Dominance Analysis}
\label{app:dominance}

Table~\ref{tab:dominance} reports how far the Consensus-based MAD outcome falls below the arithmetic mean of the Strict-only and Lenient-only score shifts, the midpoint expected under simple averaging.

\begin{table}[h]
\centering
\small
\setlength{\tabcolsep}{3pt}
\begin{tabular}{@{}lcc@{}}
\toprule
\textbf{Comparison} & \textbf{Korean Essay} & \textbf{SummEval} \\
\midrule
Mean of Strict and Lenient-only & $-0.334$ & $-0.335$ \\
Consensus-based MAD & $-0.589$ & $-0.454$ \\
\bottomrule
\end{tabular}
\caption{Aggregate-level dominance analysis, averaged across six judge models.}
\label{tab:dominance}
\end{table}

These values establish strict-stance dominance beyond averaging as the central finding: the consensus mean falls well beyond the arithmetic midpoint of the two standalone conditions.

\end{document}